\documentclass[11pt,a4paper]{article}
\usepackage{times,latexsym}
\usepackage{url}
\usepackage[T1]{fontenc}

\usepackage{acl2023}
\usepackage{booktabs}

\usepackage{xspace,mfirstuc,tabulary}

\usepackage{amsmath} 
\usepackage[disable]{todonotes}   

\newif\iftaclinstructions
\taclinstructionsfalse 
\iftaclinstructions
\renewcommand{\confidential}{}
\renewcommand{\anonsubtext}{(No author info supplied here, for consistency with
TACL-submission anonymization requirements)}
\newcommand{\instr}
\fi

\usepackage[disable]{todonotes}   

\title{Neural Machine Translation for the Indigenous Languages of the Americas: An Introduction.}

\author{ \textbf{Manuel Mager${ }^{\heartsuit}$\thanks{~~Work done while at  the University of Stuttgart.}\quad
Rajat Bhatnagar${ }^{\spadesuit}$ \quad \
Graham Neubig${ }^{\sharp}$}\\
\textbf{
Ngoc Thang Vu${ }^{\diamondsuit}$  
Katharina Kann${ }^{\spadesuit}$} \\
${ }^{\heartsuit}$AWS AI Labs \quad  \quad
${ }^{\sharp}$Carnegie Mellon University \\
${ }^{\spadesuit}$University of Colorado Boulder \quad 
${ }^{\diamondsuit}$University of Stuttgart}
\date{}
 
\begin{document}
\maketitle

\begin{abstract}
Neural models have drastically advanced state of the art for machine translation (MT) between high-resource languages.
Traditionally, these models rely on large amounts of training data, but many language pairs lack these resources.
However, an important part of the languages in the world do not have this amount of data. Most languages from the Americas are among them, having a limited amount of parallel and monolingual data, if any.   Here, we present an introduction to the interested reader to the basic challenges, concepts, and techniques that involve the creation of MT systems for these languages. Finally, we discuss the recent advances and findings and open questions, product of an increased interest of the NLP community in these languages. 

\end{abstract}

\section{Introduction}

More than 7 billion people on Earth communicate in nearly 7000 different languages \cite{pereltsvaig2020languages}. Of these, approximately 900 languages are native of the American continent \cite{campbell2000american}. Most of these indigenous languages of the Americas (ILA) are endangered at some degree \cite{thomason2015endangered}.
This huge variety in languages is simultaneously a rich treasure for humanity and also a barrier to communication among people from different backgrounds.
Human translators have been important in overcoming language barriers.
However, trained translators are not accessible to everyone on Earth and even scarcer for endangered and minority languages. The need for translations is even written in the constitutions of several countries like Mexico, Peru, Paraguay, Venezuela, and Bolivia \cite{zajicova2017lenguas} to allow native speakers to have equal language rights regarding law.

This is why developing MT is crucial: it helps humanity overcome language barriers while simultaneously allowing people to continue using their native tongue. However, the challenges to achieving these problems are not trivial. It is not only the amount of available data (a common thesis among the NLP community) but also a set of challenging issues (dialectical and orthographic variations, noisy texts, complex morphology, etc.) that must be addressed.

MT has always been an important task within the larger area of natural language processing (NLP). In 1954, the Georgetown--IBM experiment \cite{hutchins2004georgetown} was the first that showed at least some effectiveness of MT. Further research resulted in rule-based systems and statistical models.
In 2023, neural models  
define state of the art for MT if training data is plentiful -- i.e., for so-called high-resource languages (HRLs) -- and have also achieved impressive results for low-resource languages (LRLs). MT is also the most studied NLP task for the ILA \cite{mager-etal-2018-challenges,littell-etal-2018-indigenous}. The common issue among these languages is the extreme low-resource conditions they are confronted with. The research interest for these languages has increased in the last years, including the recent AmericasNLP 2021 shared task \cite{mager-etal-2021-findings} on 10 indigenous languages to Spanish, and the WMT (Conference on Machine Translation) shared task for Inuktitut--English \cite{barrault-etal-2020-findings}.

In this work we aim to provide a comprehensive introduction to the challenges that involve creating MT systems for ILA, and the current status of the existing work.
We organize this work as follows:  
We start by introducing state-of-the-art NMT models (\S\ref{sec:background}). Then, we discuss the current challenges for these languages (\S\ref{sec:challenges}); and we introduce the key concepts related to low-resource NMT and the implications for endangered languages of the Americas(\S\ref{sec:challenges}).
This is followed by a discussion of available data
(\S\ref{sec:datasets}). Afterwards, we introduce the important concepts for LRL and endangered languages (\S\ref{sec:LRMT}); then we introduce the main strategies aimed at improving NMT with limited training data (\S\ref{sec:paradigms}); and finally we give an overview of the work done for ILA on MT (\S\ref{sec:mtadvances}).
In doing so, we provide insights into the following questions: Which systems define the state of the art on low-resource NMT applied to the ILA? What is the route that ahead to improve the translations of the ILA? 

\section{Background and Definitions} 
\label{sec:background}

Formally, the task of MT consists of converting text $X$ in a source language $L_x$ into text $Y$ in a target language $L_y$ that conveys the same meaning.%
\footnote{This is an approximation, since it is in general not possible to map the meaning of text exactly into another language \cite{nida1945linguistics,sechrest1972problems,baker2018other}.}
Translating text $X \in L_x$ into $Y \in L_y$ can be described as a function \cite{neubig2017neural}:
\begin{align}
    Y = \texttt{MT}(X).
\end{align}
$X$ and $Y$ can be of variable length, such as phrases, sentences, or even documents.

If other languages are used during the translation process, e.g., as pivots, we denote them as $L_1, \dots, L_n$. We refer to a corpus of monolingual sentences in language $L_i$ as $M^{L_i} = S_1, ..., S_n$. 

\paragraph{Probabilistic Modeling and Data}
When using probabilistic MT models, the goal is to find $Y \in L_y$ with the highest conditional probability, given $X \in L_x$.
Under the supervised machine learning paradigm, a parallel corpus $C_{parallel} = (X_1, Y_1), ..., (X_n, Y_n)$ is used to learn a set of parameters $\theta$, which define a probability distribution over possible translations.  
Given $C_{parallel}$, the training objective of an NMT  model is generally to maximize the log-likelihood $\mathcal{L}$ with respect to $\theta$:
   \begin{align}
    \mathcal{L}_\theta = \sum_{(X_i,Y_i) \in C_{parallel}} \log p(Y_i|X_i;\theta).
\end{align}

Within this overall framework, there are a number of design decisions one has to make, such as which 
model architecture to use,
how to generate translations, and how to evaluate.

\paragraph{Decoding } 
Decoding refers to the generation of output $\hat{Y}$, given the parameters $\theta$ and an input $X$.
Often, decoding is done by approximately solving the following maximization problem:
\begin{align}
    \text{argmax}_{\hat{Y}}p(\hat{Y}|X;\theta) 
\end{align}
Most NMT systems factorize the probability of $\hat{Y}=\hat{y_1}, ..., \hat{y_T}$ in a left-to-right fashion:
\begin{align}
    p(\hat{Y}) = \prod^{T}_{t=1} p(\hat{y}_t | \hat{y}_{<t},X,\theta)
\end{align}
Thus, the probability of token $\hat{y}_t$ at time step $t$ is computed using the previously generated tokens $\hat{y}_{<t}$, the source sentence $X$ and the model parameters $\theta$. 
Common algorithms for finding a high-probability translation are greedy decoding, i.e., picking the token with the highest probability at each time step, and beam search \cite{lowerre1976harpy}.

\subsection{Input Representations}
\label{subsec:input_rep}
The texts $X$ and $Y$ are input into an NMT system as sequences of continuous vectors. However, defining which units should be represented as such vectors is non-trivial. 
The classic way is to represent each \emph{word} within $X$ and $Y$ as a vector (or embedding).
However, in a low-resource setting, often not all vocabulary items appear in the training data \cite{jean-etal-2015-using,luong-etal-2015-addressing}.
This issue especially effects languages with a rich inflectional morphology \cite{sennrich-etal-2016-neural}: as many word forms can represent the same lemma, the vocabulary coverage decreases drastically.
Furthermore, for many LRLs, boundaries between words or morphemes are not easy to obtain or not well defined in the case of languages without a standard orthography.
Alternative input units have been explored, such as characters \cite{ling2015character}, 
byte pair encoding \citep[BPE;][]{sennrich-etal-2016-improving}, morphological representations \cite{vania-lopez-2017-characters,ataman2018compositional}, syllables \cite{zhang2019open}, or,  recently, a visual representation of rendered text 
\cite{salesky2021robust}. No clear advantage has been discovered for using morphological segmentations over BPEs when testing them on LRLs \cite{saleva-lignos-2021-effectiveness}.

Input representations can be pretrained. The two most common options are: i) word embeddings, where each type is represented by a vector, e.g., Word2Vec \cite{mikolov2013efficient}, Glove \cite{pennington-etal-2014-glove}, or Fasttext \cite{bojanowski2017enriching}) embeddings, and 
ii) contextualized word representations, where 
entire sentences are being encoded at a time, e.g., ELMo \cite{peters2018deep} 
or BERT \cite{devlin2019bert}.
However, training of these methods requires large monolingual training corpora, which may not be readily available for LRLs. 
As most ILA have rich morphology, this topic has gathered special interest. The discussion about the usage of morpholigical segmented input for NMT models is recurrent. \cite{mager-etal-2022-bpe} show that the unsupervised morphologically inspired models outperform BPE pre-processing (experimented on 4 language pares). Similar experiments done on Quechua--Spanish and Inuktitut--Enlgish \cite{schwartz2020neural}, comparing BPEs against Morfessor \cite{smit2014morfessor}. Also \cite{ortega2020neural} improves the SOTA (state-of-the-art) for Quechua--Spanish MT using a morphological guided BPE algorithm.

\subsection{Architectures}
NMT models typically are sequence-to-sequence models.
They encode a variable-length sequence into a vector or matrix representation, which they then decode back into a variable-length sequence \cite{cho2014learning}. The two most frequent architectures are: i) recurrent neural networks (RNN), such as LSTMs \cite{hochreiter1997long} or GRUs \cite{cho2014learning}, and 
 ii) transformers \cite{vaswani2017attention}, which define the current state of the art in the high-resource setting.

As for most neural network models, training an NMT system 
on a limited number of instances 
is challenging \cite{fernandez2014we}.
There are common problems that arise from limited data in the training set. One major advantage of neural models is their ability to learn representations from raw data, in contrast to manually engineered features \cite{barron1993universal}.  However, problems arise when not enough data is provided to enable effective learning of features. Another strength of neural networks is their generalization capacity \cite{kawaguchi2017generalization}. However, training a neural network on a small dataset easily leads to overfitting \cite{rolnick2017deep}. Recent studies, however, show empirically that this does not necessarily happen if the network is tuned correctly \cite{olson2018modern}.

\subsection{Evaluation}

Accurately judging translation quality is difficult and, thus, often still done manually:
 bilingual speakers 
assign scores according to provided criteria such as fluency and adequacy (\emph{Does the output have the same meaning as the input?}). However, manual evaluation is expensive and slow. Moreover, in the case of endangered languages, bilingual speakers can be hard or impossible to find.

Automatic metrics provide an alternative.\footnote{For a detailed overview of automatic metrics for MT we refer the interested reader to specialized reviews \cite{han2016machine,celikyilmaz2020evaluation,chatzikoumi2020evaluate}.} These metrics assign a score to system output, given one or more ground truth reference translations. 
The most widely used metric is BLEU \cite{papineni-etal-2002-bleu}, which relies on token-level $n$-gram matches between the translation to be rated and one or more gold-standard translations.
For morphologically rich languages, character-level metrics, such as chrF \cite{popovic2017chrf++}, are often more suitable, as they allow for more flexibility. In the AmericasNLP ST \cite{mager-etal-2021-findings} this metric was used over BLEU, as it fits better to the rich morphology of many ILA.

To have a concrete example, lets have the following Wixarika phrase with an English translation:
\begin{center}

  \begin{tabular}{ l  l }
yu-huta-me & ne-p+-we-'iwa \\
an-two-ns & 1sg:s-asi-2pl:o-brother 
  \end{tabular}
  
\emph{I have two brothers} 
\end{center}

As discussed in \cite{mager-etal-2018-lost} it is difficult to translate back from Spanish (or other Fusional language) the morpheme \textit{p+} as it has not equivalent in these languages. So if we would ignore these morpheme at all, BLEU would penalize the entire word \textit{nep+we'iwa}. In contrast, chrF would give credit to the translation, even if the \textit{p+} is missing.

One  shortcoming of these evaluation metrics is that the evaluation is very dependent on the surface forms and not on the ultimate goal of semantic similarity and fluency.
Recent work uses pretrained models to evaluate semantic similarity between translations and the gold standard \cite{zhang2020bertscore}, but these methods are limited to languages for which such models are available. This is not possible for the ILA, as the amount of monolingual data is not enough to train a reliable pretrained language model\footnote{One exception to this is Quechua, that has a large enough monolingual dataset to train a BERT like model \cite{zevallos-etal-2022-introducing}}.

\section{Challenges and open questions}
\label{sec:challenges}

In an overview of the datasets and recent studies of MT for the ILA, we found the following main issues to be handled.

\paragraph{Extreme low-resource parallel datasets} 
Even with the recent advances, the resources available to 
train MT systems are extremely scarce, having training set between 4k and 20k sentences (see \S\ref{sec:datasets}), 
with notable exceptions for Inuktitut, Guarani and Quechua \cite{joanis-etal-2020-nunavut,ortega2020neural}.

\paragraph{Lack of monolingual data} Most of these languages are mostly used in spoken form. In recent years, with the advancement and democratization of mobile technologies, indigenous languages have seen a slight increase in massaging systems and private spheres \cite{rosales2019towards}. However, the usage of these languages on the internet is rather limited. Even Wikipedia has a limited amount of these languages \cite{mager-etal-2018-challenges}.

\paragraph{Low domain diversity}. As most parallel datasets are scarce, they are restricted to a small number of domains, making it challenging to adapt it, or try to aim for general translation models. This has been recognized as a major problem during the AmericasNLP ST \cite{mager-etal-2021-findings}. 

\paragraph{Rich morphology} An important number of these languages are morphological highly rich. In many cases, we find polysynthetic, with or highly agglutinative languages \cite{kann-etal-2018-fortification} or even fusional phenomenon \cite{mager2020tackling}. 

\paragraph{Distant paired language} The most common languages that we find that ILA is 
translated into are Spanish, English, and Portuguese. However, these languages are distantly related to 
the ILA, and have completely different linguistically phenomenons \cite{campbell2000american, romero2016richard}. 

\paragraph{Noisy text environments} Monolingual texts, if exist, are found in social media that often use a non-canonical witting \cite{rosales2019towards}.

\begin{table*}[h!]
    \centering
    \small
    \setlength{\tabcolsep}{2.5pt}

    \begin{tabular}{l p{0.3\linewidth} lll}
        \toprule
        \textbf{Dataset} & \textbf{Paired-languages}& Authors \\ \midrule
        AmericasNLI & Aymara, Asháninka, Bribri, Guaraní, Nahuatl, Otomí, Quechua, Rarámuri, Shipibo-Konibo, Wixarika & \cite{ebrahimi-etal-2022-americasnli}\\
        CPML & Ch’ol, Maya, Mazatec, Mixtec, Nahuatl and Otomi  &  \cite{sierra-martinez-etal-2020-cplm}\\
        OPUS & * &  \cite{tiedemann2016opus} \\
        New testament Bible & * &  \cite{mccarthy-EtAl:2020:LREC1}\\
    \bottomrule
    \end{tabular}
    \caption{Parallel dataset collections that contain one or more indigenous languages of the Americas}
    \label{tab:languages_all}
\end{table*}
\paragraph{Code-Swithing} This phenomenon is strongly present in ILA, as all of these languages are minority languages in their own countries. The bilingualism among their communities is strong (and CS is a common phenomenon in this setup \cite{cetinoglu-2017-code}). The final result of this phenomenon is the inclusion of code-switching on a common base \cite{mager-etal-2019-subword} in their language. 

\paragraph{Lack of orthographic normalization} The usage of ILA faces the problem of having a unified orthographic standard. This is not always possible, as the suggestions of linguists and official entities do not always match the day-by-day writing of the speakers. Moreover, in some cases, special symbols present in the orthographic standards are not accessible in English or Spanish keyboard and need to be replaced with other symbols. The winner of the AmericasNLP ST got important improvements using orthographic normalizers developed specifically for each American language \cite{vazquez-etal-2021-helsinki}.

\paragraph{Dialectal variety} The indigenous languages have a strong dialectal variety, making it hard for native speakers to understand even speakers from neighboring villages. The linguistic richness of entire regions is so diverse that even a single state like the Mexican Oaxaca could correspond to the diversity in the whole Europe \cite{mcquown1955indigenous}. 

\section{Available MT datasets for ILA}
\label{sec:datasets}

\begin{table*}[]
    \centering
    \small
    \setlength{\tabcolsep}{2.5pt}

    \begin{tabular}{llllrrr p{1.5cm} p{3cm}}
        \toprule
        \textbf{Language} & \textbf{Paried-language}&\textbf{ISO} & \textbf{Family} & \bf Sentences & & & Domain  &Authors \\ \midrule
        Asháninka & Spanish & cni & Arawak & 3883 &  &  &  & \cite{ortega-etal-2020-overcoming} \\
        Bribri & Spanish &bzd & Chibchan & 5923 &  &  &  & \cite{feldman-coto-solano-2020-neural} \\
        Guarani & Spanish &gn & Tupi-Guarani & &  &  & News, Blogs & \cite{abdelali-etal-2006-guarani}\\
        Guarani & Spanish &gn & Tupi-Guarani & 14,531 &  &  & News, Blogs & \cite{chiruzzo-etal-2020-development}\\
        Guarani & Spanish &gn & Tupi-Guarani & 14,792 &  &  & News, Social Media & \cite{gongora-etal-2021-experiments}\\
        Guarani & Spanish & gn & Tupi-Guarani & 30855 & && 8 Domains & \cite{chiruzzo2022jojajovai}\\
        Nahuatl & Spanish &nah & Uto-Aztecan & 16145 &  &  & Diverse Books & \cite{gutierrez-vasques-etal-2016-axolotl}\\
        Otomí & Spanish &oto & Oto-Manguean  & 4889 &  &  & Diverse Books & \url{https://tsunkua.elotl.mx}\\
        Rarámuri & Spanish &tar & Uto-Aztecan & 14721 &  &  & Dictionary Examples & \cite{mager-etal-2022-bpe} \\
        Shipibo-Konibo & Spanish &shp & Panoan & 14592 &  &  & Educational, Religious & \cite{galarreta-etal-2017-corpus}\\
        Wixarika & Spanish &hch & Uto-Aztecan & 8966 &  &  & Literature & \cite{mager2018probabilistic}\\
        Cherokee & English & chr & Uto-Aztecan &  &  &  & OPUS& \cite{zhang-etal-2020-chren}\\
        Inuktitut & English & iku & Eskimo–Aleut  & 1,450,094 &  &  & Legislative & \cite{joanis-etal-2020-nunavut}\\
        Ayuuk & Spanish & mir &  Mixe–Zoque & 7553 &  &  & Diverse & \cite{zacarias-marquez-meza-ruiz-2021-ayuuk}\\
        Mazatec& Spanish & Many &  Oto-Manguean & 9799  &  &  & Diverse & \cite{tonja2023parallel}\\
        Mixtec & Spanish & Many &  Oto-Manguean & 13235   &  &  & Diverse & \cite{tonja2023parallel}\\
        
    \bottomrule
    \end{tabular}
    \caption{Parallel datasets that have been released focusing on one indigenous language}
    \label{tab:languages_specific}
\end{table*}

The parallel datasets available for MT have been increasing during the last years. At this moment, we can show in two folds the development of these resources: as shown in table \ref{tab:languages_specific} work on specific language has emerged; but also broader datasets have started to cover the ILA (see table \ref{tab:languages_all}). 

Language-specific corpus collection work has been done for many languages, where parallel corpus has been the main component. In recent time we have seen  Cherokee--English (OPUS) \cite{zhang-etal-2020-chren}, Wixarika--Spanish \cite{mager2018probabilistic}, Shipio--Konibo \cite{feldman-coto-solano-2020-neural}, and others (see table \ref{tab:languages_specific}). The most prominent of these datasets has been the Inuktitut--English parallel data. 
The last version of this dataset corpora \cite{joanis-etal-2020-nunavut} is has medium size with  1,450,094 sentences. Previous versions of this corpus are \cite{martin2003aligning}. This data set was used for the WMT 2020 Shared Task on Unsupervised, and Low Resourced MT \cite{barrault-etal-2020-findings}.

For wide-spoken languages like Guarani, it is even possible to collect a web crawled dataset, including news articles and social media parallel aligned data \cite{chiruzzo-etal-2020-development,gongora-etal-2021-experiments} This dataset also includes monolingual data. This is possible as Guaraní is one of the most spoken indigenous languages of the continent. 

In contrast to the language-specific datasets, we find broader approaches (see table \ref{tab:languages_all}). The broadest multilingual dataset, which contains the Bible's New Testament, includes about 1600 languages  \cite{mayer2014creating,mccarthy-EtAl:2020:LREC1} of
 the 2,508 that have been collected by the Summer Institute of Linguistic (SIL) \cite{anderson2012languages}. Another remarkable effort to obtain broad language coverage is the PanLex project \cite{kamholz2014panlex}, which has gathered lexical translation dictionaries for over 5,700 languages. However, for most languages, PanLex contains only a few dozen words. \newcite{duan2020bilingual} show that such dictionaries can be used to create an NMT system, making bilingual dictionaries relevant for further studies. 

Recently community-driven research groups have started the creation of own parallel datasets, such as Masakhane \cite{orife2020masakhane,nekoto-etal-2020-participatory} for African languages, and AmericasNLP for indigenous languages of the Americas \cite{ebrahimi2021americasnli,mager-etal-2021-findings}. The AmericasNLI dataset is an important effort to have a common evaluation benchmark for the 10 indigenous languages of the Americas for the MT and NLI tasks.

Given the constitutional rights of indigenous languages in many countries of the Americas, it is possible to access this data. \newcite{vazquez-etal-2021-helsinki} made available this resource during their shared task system development. 

Finally, it is important to mention that many of the languages spoken in the Americas have Wikipedia's set of articles available\footnote{The available languages in wikipedia can be consulted at: \url{https://es.wikipedia.org/wiki/Portal:Lenguas_indígenas_de_América}. Until the publication of this article, there were only entries in Nahuatl, Navajo, Guarani, Aymara, Klaalisut, Esquimal, Inukitut, Cherokee, and Cree.}.

\paragraph{Collection of New Data }
A common way to create parallel data with the help of bilingual speakers is via elicitation (translating the foreign text into another language). It has the disadvantage of biasing the created text to forms and topics, culture, and even grammatical forms towards the source language \cite{lorscher2005translation}. A method that avoids this problem is language documentation, which consists of storing and annotating commonly used speech or text \cite{himmelmann2008language}. However, it is costly and requires specialists. In this process, involving the community members that are bilingual speakers is important \cite{bird-2020-decolonising}. 

\section{Low-resource MT}
\label{sec:LRMT}

\begin{figure*}
    \centering
    \includegraphics[width=.73\linewidth]{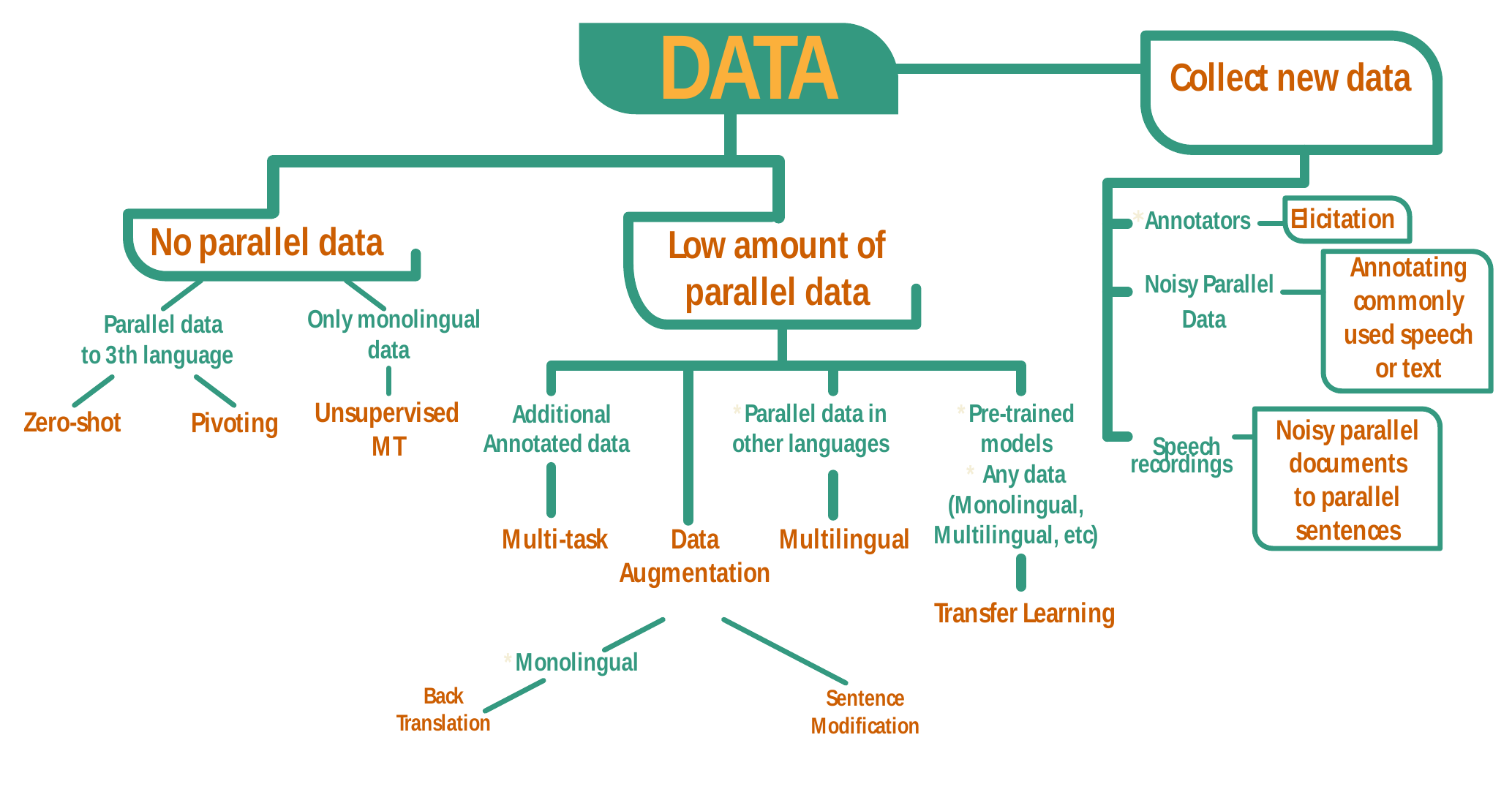}
    \caption{What to do when we have low o no data to train our machine translation models? This diagram shows basic scenarios, solutions, and common requirements for each method, with the section describing the method.}
    \label{fig:low_resouce_mt}
\end{figure*}

For the purpose of this paper we define LRLs as languages 
for which standard techniques are unable to create
well performing systems,  
which makes it necessary to resort to other techniques (cf. Figure \ref{fig:low_resouce_mt})
such as transfer learning.
For MT, the amount of available resources differs widely across language pairs: some have less than 10k parallel sentences, while other have more than 500k, with some exceptions in the orders of several million.

Emulating a low-resource scenario by down-sampling 
available data for high-resource languages is common and helps understanding a model's performance across different settings. 
However, further evaluating methods on a diverse set of low-resource languages is crucial, 
since many languages exhibit particular linguistic phenomena \cite{mager2020tackling}, that perturb the final results, 
especially since most large datasets are from the Indo-European language family, to which only 6.16\% of the world's languages belong \cite{lewis2009ethnologue}.

Importantly, there is no strong correlation between the number of resources available per language and the number of speakers: Javanese with 95 million speakers and Kannada with 44 million are considered LRLs, while French, with only 64 million native speakers, is among the most widely studied languages. Improving models to handle LRLs will
extend access to information online as well as human language technology to all monolingual speakers of those languages. In the case of ILA, most languages are endangered at some degree, but most of them have the same issue: they are low resourced for parallel and monolingual data.

\paragraph{Endangered Languages } \newcite{krauss1992world} estimates 
that 50\% of all languages are doomed or dying, and that in this century we will see either the death or the doom of  90\% of all human languages.  The current proportion of languages that are already extinct or moribund ranges from 31\% down to 8\% depending on the region, with the most severe cases in the Americas and Australia \cite{simons2013world}. 
To determine how endangered a language is, \newcite{lewis2010assessing} proposes a classification scale called EGIDS with 13 levels. The higher the number on this scale, the greater the level of disruption of the language's inter-generational transmission.\footnote{The complete EGIDS scale can be found at \url{https://www.ethnologue.com/about/language-status}} 
 MT for endangered LRLs has the potential to help with documentation, promotion and revitalization efforts \cite{galla2016indigenous,mager-etal-2018-challenges}. 
 However, as these languages are commonly spoken by small communities, or indigenous people, researchers should aim for a direct involvement of those communities \cite{bird-2020-decolonising}.


\paragraph{What is polysynthesis?}
\label{sec:poly}
A polysynthetic language is defined by the following linguistic features: 
the verb in a polysynthetic language must have an agreement with the subject, objects and indirect objects \cite{baker1996polysynthesis};
nouns can be incorporated into the complex verb morphology \cite{mithun1986nature}; and, therefore, polysynthetic languages have agreement morphemes, pronominal affixes and incorporated roots in the verb \cite{baker1996polysynthesis}, and also encode their relations and characterizations into that verb. The most common word orders present in these languages are SOV, VSO, SVO and free order. It is important to notice that a polysynthtic language can have a aggutinative \footnote{Agglutination refers to a concatenation of morphemes, with minimal changes to the surface form.} or can have also fusional characteristics, like Totonaco or Tepehua \cite{mager2020tackling}.

\section{Low-resource MT paradigms}
\label{sec:paradigms}

Most languages of the Americas do not have high amount of data for MT. Therefore, we introduce the most important paradigms to improve low-resourced machine translation. Figure \ref{fig:low_resouce_mt} shows a general overview of the methods and options to improve LRL MT. For a more detailed understanding of this techniques we refer the reader to specialized low-resource MT surveys \cite{haddow2022survey,wang2021survey,ranathunga2021neural}.

\subsection{Multilingual Supervised Training}
\label{sec:mutlilingual_sup}

\begin{figure}[t]
    \centering
    \includegraphics[width=.40\columnwidth]{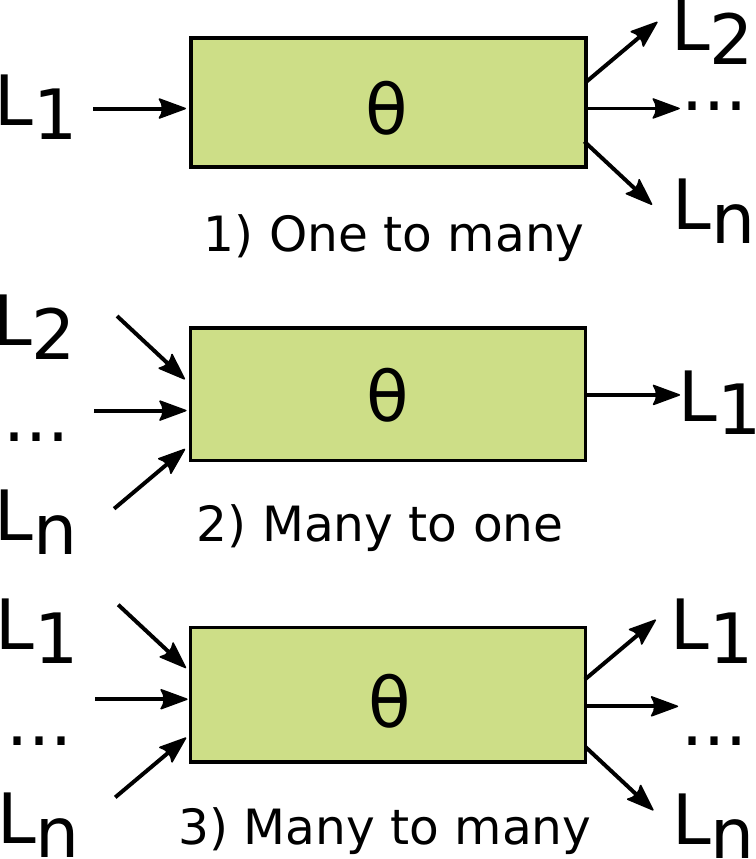}
    \caption{An overview of different multilingual setups.}
    \label{fig:multilingual}
\end{figure}
With a multilingual set of parallel data $D_{parallel}$ between different language pairs $\{(L_1,L_2), \dots, (L_m,L_n)\}$ we can train a model that is able to map a sentence from any source language $L_x$ into any target language $L_y$ that is contained in $D_{parallel}$ (see \ref{fig:multilingual}). These multilingual NMT models have seen a growth in popularity and efficiency in recent years. We will now cover the different training algorithms for these models: 
1) many source languages and one target language (\textit{many-to-one}), 2) one source and many target languages (\textit{one-to-many}), and 3) many source languages and many target languages (\textit{many-to-many}). For a general overview of multilingual MT, we refer the reader to  surveys dedicated to this topic \cite{tan2019study,dabre2019survey}.
\newcite{johnson2017google} are the first to introduce a multilingual NMT model, trained on translating from a large number of languages to English as well as in the opposite direction. The authors show that these models improve over single-language pair models for LRLs.

\subsection{Multi-task Training}
\label{sec:multi_task}
Multi-task training \cite{caruana1997multitask} aims to improve the performance of the main task -- MT in our case -- by adding one or more auxiliary tasks to the training.  
The easiest way is to share all parameters of the network, using the ideas already explored in multilingual NMT (\S\ref{sec:mutlilingual_sup}). This can be done with a special flag in the input that specifies the current task. It is also possible to share only the encoder and have two separate decoders for each task. 

\paragraph{Multilingual Modeling}
\label{sec:multilingual_mod}
In order to handle multilinguality it is also possible to adapt modify the NMT models. The main proposals to do so has been: sharing all parameter except the attention mechanism of a RNN NMT model \cite{blackwood2018multilingual}; parameter sharing in the transformer architecture \newcite{sachan2018parameter};

\subsection{Data Augmentation}
\label{sec:data_augmentation}

\paragraph{Back-Translation}
\label{subsec:backtranslation}

\begin{figure}
    \centering
    \includegraphics[width=.60\columnwidth]{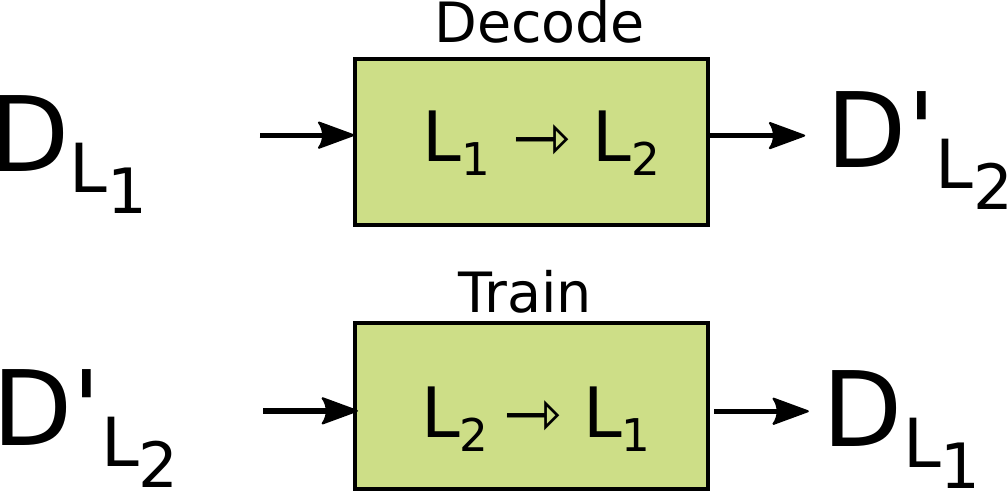}
    \caption{Backtranslation}
    \label{fig:backtranslation}
\end{figure}

A straightforward way to leverage monolingual data for low-resource MT is to generate a meaningful signal with the help of an already initialized MT model (see Figure \ref{fig:backtranslation}). This method is called back-translation \citep[BT;][]{sennrich2016improving}: 
With monolingual data $M^{L_{x}}$ in source language $L_x$ and a trained model that is able to translate from $L_x$ into a target language $L_y$ we can generate a translation $M'^{L_{y}}$. This pseudo parallel data $(M^{L_{x}}, M'^{L_{y}})$ is then used to train a new model in the opposite direction. This process can be applied iteratively to improve the translation \cite{hoang2018iterative}.

\paragraph{Sentence Modification}
Other methods to generate more parallel sentences are based on lexical substitution. \newcite{fadaee2017data} explores replacing frequent words with low-frequency ones in both source and target to improve the translation of rare words. This is done using language models (LMs) and automatic alignment.

\paragraph{Pivoting}
\label{subsubsec:pivoting}
If no parallel corpus between languages $L_x$ and $L_y$ is available, but both of them have parallel corpora with a third language $L_p$, pivoting is an option. The basic idea is to train two MT systems: one that translates $L_x \to L_p$ and another for $L_p \to L_y$. Pivoting has first been introduced for SMT \cite{wu2007pivot,cohn2007machine,utiyama2007comparison}.

\subsection{Semi-supervised and Unsupervised MT}
\label{sec:semi_and_unsupervised}

\paragraph{Transfer Learning via Pretraining}
\label{subsec:transfer_learning}
Transfer learning refers to using knowledge learned from one task to improve performance on a related task \cite{weiss2016survey}. In recent years this approach has gained popularity with big multilingual models such as \newcite{conneau2019cross} that proposes training the encoder and the decoder separately in order to get cross-language representations (XLM). This idea has further been extended by \citet[][ MASS]{song2019mass} to masking a \textit{sequence} of tokens from the input (multilingual MASS \cite{siddhant-etal-2020-leveraging}).
Another approach is to train the entire transformer model as a denoising autoencoder \citep[BART; ][]{lewis2019bart} (
multilingual BART (mBART)  \cite{liu2020multilingual}). It is also possible to pretrain a transformer in a multi-task, text-to-text fashion, where one of the tasks is MT \citep[T5;][]{raffel2020exploring} (multilingual version \cite{xue2021mt5}).

\paragraph{Unsupervised MT}
\label{subsec:umt}
UMT covers approaches that do \textit{not} require any parallel text, relying only on monolingual data. This differs from zero-shot translation, which uses parallel data for other language pairs.
Early approaches tackled the problem with an auto-encoder with adversarial training \cite{lample2017unsupervised} or with auto-encoders with a shared encoding space as well as separate decoders for each target language \cite{artetxe2018unsupervised}. The main problem for these approches is the need of a big monolingual dataset, that is not available for most ILA.

\section{Advances in MT for the indigenous languages of the Americas}
\label{sec:mtadvances}

In recent years the interest in MT for indigenous languages of the Americas has increased. 
The task is not easy. The first usage of NMT systems has not been successful \cite{mager2021retos}. However, with the use of LRL MT methods, we have witnessed great improvements.

The Cherokee--English \cite{zhang-etal-2020-chren} language pair has been explored using a pre-trained BERT \cite{devlin2019bert} for the English side. A system demonstration of this approach is also accessible \cite{zhang-etal-2021-chrentranslate}. 
The back translation strategy for Bribri--Spanish NMT transformers has also been explored \cite{feldman-coto-solano-2020-neural} and by \cite{oncevay-2021-peru} (for four Peruvian languages to Spanish) with good results. The scarce indigenous language monolingual text can be replaced to some extent with Spanish text or extracted from PDFs, and other sources \cite{bustamante-etal-2020-data}.

One of the main challenges for the complex morphological languages in the area has been the prepossessing step. \newcite{schwartz2020neural} show that even if morphological segmentation has less perplexity a the language modeling time, it is still under-performing or equivalent against BPEs for MT (for Inuktitut-–English, Yupik–-English Data, Guaraní–-Spanish Data). A more comprehensive (on the segmentation modeling side) was done by \cite{mager-etal-2022-bpe} exploring a wide array of segmentation models.
The latter study showed that supervised morphological segmentation under-perform unsupervised. However, unsupervised morphological segmentation like LMVR \cite{ataman2017linguistically} and FlatCat \cite{gronroos2014morfessor} perform better than BPEs. \cite{ngoc-le-sadat-2020-revitalization} studied how better to perform word segmentation for the Inuktitut--English pair. They found that for this language pair, a morphological segmentation, or a combination of BPEs and morphological segmentation, works better than just applying vanilla BPEs.
Also, training word embeddings for Guarani--Spanish translation is an excellent opportunity to increase the MT performance of these languages \cite{gongora-etal-2022-use}.

The usage of transfer learning from multilingual systems has been tried, with limited results \cite{nagoudi-etal-2021-indt5} (training an own T5 model for indigenous languages) and \cite{zheng-etal-2021-low}. However, pertaining a Spanish--English model together with ILA, and then fine-tuning it (together with a careful prepossessing and filtering step) has been the most successful strategy \cite{vazquez-etal-2021-helsinki}.

The quality of MT systems of ILA has been a constant debate. However, \newcite{ebrahimi2021americasnli} shows that the quality of MT for these languages is enough to improve other tasks like natural language inference (NLI).

\paragraph{Inuktitut--Enlgish ST}

The WMT 2020 news translation task included Inuktitut--English translation \cite{barrault-etal-2020-findings}. The participating systems explored the difficulties of working with a polysynthetic language in a medium resource scenario. 
Participating teams in this competition were: \cite{kocmi-2020-cuni,hernandez-nguyen-2020-ubiqus,scherrer-etal-2020-university,roest-etal-2020-machine,lo-2020-extended,knowles-etal-2020-nrc,zhang-etal-2020-niutrans,krubinski-etal-2020-samsung}.

\paragraph{AmericasNLP 2021 and 2023 ST } In 2021, the AmericasNLP community organized a workshop on Machine Translation for 10 indigenous languages of the Americas in 2021 \cite{mager-etal-2021-findings} and 2023 \cite{ebrahimi-etal-2023-findings} with an additional indigenous language (Chatino). The AmericasNLP shared task winner was \cite{vazquez-etal-2021-helsinki} in 2021, and a more mixed result in 2023\footnote{Up to this moment, no official desciption papers for the 2023 are published.}. Other participants in this shared task are \cite{nagoudi-etal-2021-indt5,bollmann-etal-2021-moses,zheng-etal-2021-low,knowles-etal-2021-nrc,parida-etal-2021-open,nagoudi-etal-2021-indt5}. It is important to point at the importance of clean datata. For Quechua, \cite{moreno-2021-repu} got the best results generating an additional amount of clean data.

\paragraph{AmericasNLP 2022 Competition} is a competition on Speech-to-Text translation is organized and is targeting the following language pairs: Bribri–Spanish, Guaraní–Spanish, Kotiria–Portuguese, Wa'ikhana–Portuguese, and Quechua–Spanish \cite{ebrahim-et-al-2022-findings}\footnote{\url{http://turing.iimas.unam.mx/americasnlp/st.html}}.

\section{Ethical aspects}

When working with ILAs are also interacting with communities and nations that speak these languages. In most cases, these speakers have been exposed to a colonial past, or to a local oppression, by the majority language and culture. It is important to point to best practices and recommendations when performing our research. \newcite{bird-2020-decolonising} and \newcite{liu2022not} advocate to include community members as co-authors \cite{liu2022not} as well as considering  data and technology sovereignty. This is also aligned with the community building aimed at by \citet{zhang2022can}. \newcite{mager2023ethical} summarizes the main aspects that should be considered as follows: i) \textit{Consultation, Negotiation and Mutual Understanding}. It is important to inform the community about the planned research, negotiating a possible outcome, and reaching a mutual agreement on the directions and details of the project should happen in all cases.
ii) \textit{Respect of the local culture and involvement}. As each community has its own culture and view of the world, researchers should be familiar with the history and traditions of the community. Also, it should be recommended that local researchers, speakers, or internal governments should be involved in the project.
iii) \textit{Sharing and distribution of data and research}. The product of the research should be available for use by the community, so they can take advantage of the generated materials, like papers, books, or data.

\section{Conclusion}
\label{sec:conclusions}
Machine translation for ILA has gained interest in the NLP community over the last few years. Here, we provide an exhaustive overview of the basic MT concepts and the particular challenges for MT for ILA (in the context of low-resource scenarios and its relation to endangered languages). We additionally survey the current advances of MT for these languages.

\section*{Limitations}

This paper's aim is to give an introduction to researchers, students, of interested community indigenous community members to the topic of Machine Translation for Indigenous languages of the Americas. Therefore, this paper is not an in-depth survey of the literature on indigenous languages nor a more technical survey of low-resource machine translation. We would point the reader to more specific surveys on these aspects \cite{haddow2022survey,mager-etal-2018-challenges}.

\section*{Ethical statement}

We could not find any specific Ethical issue for this paper or potential danger. Nevertheless, we want to point to the reader that working with indigenous languages (in this case, MT) implies a set of ethical questions that are important to handle. For a deeper understanding of the matter, we suggest specialized literature to the reader \cite{mager2023ethical,bird-2020-decolonising,schwartz2022primum}.

\bibliography{tacl2018,anthology}
\bibliographystyle{acl_natbib}

\clearpage
\newpage
\appendix
\section{Appendix}

In this appendix we expand the information regarding current work on MT for LRL. 

\subsection{Expanded LR work on Multilingual supervised training}

 \newcite{arivazhagan2019missing} introduce a representational invariance training objective across languages that achieves comparable results with pivoting methods.
Promising results of multilingual models have encouraged experiments with models trained on a massive amount of language pairs, resulting in large multilingual models: \newcite{aharoni2019massively} train a single model on 102 languages to and from English in contrast to the 58 languages used by \citet{neubig2018rapid}.

The negative aspect of this approach is the size of the network. 
\citet{arivazhagan2019massively} perform an extensive study on 102 language pairs to explore different settings and training setups and achieve good results for LRLs, while maintaining good performance for high-resource languages. 
Related massively multilingual NMT systems have been trained for analytic proposes \cite{tiedemann2018emerging,malaviya2017learning} and general zero-shot transfer learning \cite{artetxe2019massively}. mRASP \cite{lin-etal-2020-pre} use for pretraining of the multilingual model and add a randomly aligned substitution loss that aims to bring words and phrases closer in the cross-lingual space.

\newcite{zhang2020improving} explores the main problems that arise for such models: multilingual NMT usually underperforms bilingual models \cite{arivazhagan2019massively}, the larger the number of languages gets the more the performance drops \cite{aharoni2019massively}, languages in datasets used for multilingual training are unbalanced in size, and poor zero-shot performance compared to pivot models (cf. \S\ref{subsubsec:pivoting}). 
\newcite{zhang2020improving} addresses these problems with a language-aware input layer, a deep transformer architecture \cite{wang-etal-2019-learning-deep}, and an online back-translation approach. These modifications boost zero-shot translation performance for multilingual models.  

To improve the problem of imbalanced and linguistically diverse training data, mostly heuristic methods have been proposed: \citet{arivazhagan2019massively} samples training data from different languages based on a data size scaled by temperature term.  These heuristics have an impact on performance, and ignore other factors that are not size. Oversampling of data is used by \newcite{johnson2017google,neubig2018rapid,conneau2019cross}. \newcite{wang-etal-2020-balancing} proposes a differentiable data selection method that automatically learns to weight training data, optimizing translation on all languages.

\paragraph{Multilingual modeling}

Sharing all parameters except for the attention mechanism shows improvements compared with sharing everything in an RNN NMT model \cite{blackwood2018multilingual}. \newcite{sachan2018parameter} explores parameter sharing in the transformer architecture for the decoder in the one-to-many translation setting and shows that transformers are more suitable than RNNs for this task. Also, parameter sharing in the decoder and embedding layer further improves performance.
\newcite{lu2018neural}  proposes a shared layer intended to capture the interlingua knowledge and an extension to the typical RNN network with multiple blocks along with a trainable routing network.  The routing network enables adaptive collaboration by dynamic sharing of blocks conditioned on the task at hand, input, and model state \cite{zaremoodi2018adaptive}.
\newcite{zhang2020improving} proposes a language-aware layer to improve such architectures further. With a similar idea, \citet{zhu-etal-2020-language} incorporates two special language embeddings into the self-attention mechanism. The first encodes the unique characteristics of each language, while the second captures common semantics across languages. 

One problem in multilingual NMT systems is the translation into the wrong language.  To address this problem, \newcite{zhang-etal-2020-improving} add a language-aware layer normalization and a linear transformation that is inserted between the encoder and the decoder to induce a language-specific translation. \newcite{raganato-etal-2021-empirical} explore to weight the target language label with jointly training  one  cross  attention  head  with  word alignments.

Other modifications of NMT model architectures to improve their performance on low-resource languages include: 
deep RNNs \cite{miceli2017deep}, normalization layers \cite{ba2016layer}, direct lexical connections \cite{nguyen-etal-2015-improving}, word embedding layers conducive to lexical sharing \cite{wang2019sde}.

\subsection{Extended Multi-task training}

\newcite{zhou-etal-2019-improving} uses this approach, but extends it with a cascade architecture: the first decoder reads the encoder, and the second decoder reads the encoder and the first decoder \cite{niehues-etal-2016-pre,anastasopoulos-chiang-2018-tied}. The auxiliary task (first decoder) is a denoising decoder. With RNN NMT architectures, one can further decide if the attention mechanism should be shared among tasks \cite{niehues-cho-2017-exploiting}. The authors compare all architectures and find that they perform similarly, with only sharing the encoder being slightly better.

Using linguistic information as an auxiliary task has not yet been explored exhaustively. \newcite{niehues-cho-2017-exploiting} studies the usage of part-of-speech (POS) and named entity (NE) tags, finding that training on named entity recognition (NER), POS tagging and MT together improves performance the most. 
 For agglutinative languages, morphological auxiliary tasks can be beneficial: \newcite{pan-etal-2020-multi} uses stemming with fully shared parameters.
 
As an alternative to linguistically informed auxiliary tasks \newcite{srinivasan2019multitask} uses multiple BPE vocabulary sizes to generate different segmentations. Each segmentation is treated as an individual task. 

\subsection{Data augmentation}

\paragraph{Back-translation}
\newcite{caswell2019tagged} shows that adding a special tag to the synthetic data improves performance. A technique that exploits this idea is training an initial translation model with synthetic data generated via BT and then finetune it with gold data \cite{abdulmumin2019tag}.
This simple yet effective training algorithm improves NMT for LRLs; however, it can also degrade performance on HRLs if trained without a tagging strategy \cite{marie-etal-2020-tagged}.

Multiple improvements of BT have been proposed. 
\newcite{edunov2018understanding} shows that sampling or noisy beam search can generate more effective pseudo-parallel data.
However, for LRLs an optimal beam search and greedy decoding are better.
A factor that influences BT's effectiveness is the quality of the initial MT systems \cite{hoang2018iterative}. 
Using back-translated data from multiple sources \cite{poncelas2019combining} or optimizing the ranking of back-translated data yields further gains \cite{soto-etal-2020-selecting}.

BT results in gains when the parallel corpora are naturally occurring text and not translationese, as the latter would only improve automatic n metrics \cite{toral2018attaining,graham-etal-2020-statistical}. 
\newcite{edunov-etal-2020-gn} 
shows that BT produces more fluent text and is preferred by humans. Additionally, translationese and original data can be modeled as separate languages in a multilingual model \cite{riley-etal-2020-translationese}. BT is also a central part of unsupervised MT (UMT; cf. \S\ref{subsec:umt}) and
zero-shot MT \cite{gu2019improved}. 

\paragraph{Sentence modification}
\newcite{zhu2019soft} proposes to replace a randomly chosen word in a sentence with a \textit{soft-word}. That means that, instead of sampling a word from the lexical distribution of a LM like \newcite{kobayashi2018contextual}, the authors use the hidden state vector of the LM directly. 
\newcite{wu2019conditional} substitutes the RNN LMs from previous work and use BERT \cite{devlin2019bert} -- a transformer trained with a masked language modeling objective -- instead. The authors finetune BERT with a conditional masked language modeling objective that tries to avoid the prediction of words that do not correspond to the original sentence meaning.

Another way to augmented MT data is by paraphrasing. If a good paraphrase system exists, this can increase the number of training instances \cite{hu2019improved}. Paraphrasing can also be used at training time by sampling paraphrases of the reference sentence from a paraphraser and training the MT model to predict the distribution of the paraphraser \cite{khayrallah2020simulated}. This helps the model to generalize. 
\citet{wieting2019beyond} propose a similar approach, using minimum risk training to optimize BLEU. To avoid BLEU's constraints to a specific reference, they use paraphrasing to diversify the given reference. 

Finally, existing data can be augmented by adding noise. This noise can be continuous or discrete. In the case of applying continuous noise, noise vectors are added to the word embeddings  \cite{cheng2018towards,sano2019effective}. Discrete noise is realized by inserting, deleting, or replacing words, BPE tokens, or characters to expand the training set in an adversarial fashion \cite{belinkov2018synthetic,ebrahimi-etal-2018-adversarial,michel2019n,cheng2019robust,cheng-etal-2020-advaug}.

\paragraph{Pivoting}
While it is simple to implement and effective, pivot-based approaches suffer from error propagation. To overcome that for NMT, joint training \citet{zheng2017maximum,cheng2019joint} and round-trip training \cite{ahmadnia2019augmenting} have been proposed.

Pivoting with NMT systems has been used for translating Japanese, Indonesian, and Malay into Vietnamese \cite{hai2019levering}, translation of related languages \cite{pourdamghani2019neighbors}, multilingual zero-shot MT \cite{lakew2018improving}, and UMT (cf. \S\ref{subsec:umt}) between distant language pairs \cite{leng2019unsupervised}.

\subsection{Recent low-resource Shared Tasks}
 First, the LoResMT 2020 shared task \cite{ojha-etal-2020-findings} explores the case of language pairs which have no parallel data between them (Hindi--Bhojpuri,  Hindi--Magahi, and Russian--Hindi). The winning system \cite{laskar-etal-2020-zero} uses a MASS model in a zero-shot fashion with additional monolingual data (see \S\ref{sec:semi_and_unsupervised}). Second, the WMT 2020 shared tasks on UMT and very low-resource supervised MT \cite{fraser-2020-findings} provide text and 60k aligned phrases for German--Upper Sorbian.,
The most important technique in all tracks is transfer learning, achieving surprisingly good results. For the AmericasNLP 2021 shared task on open MT \cite{mager-etal-2021-findings}, 10 indigenous language languages were paired with Spanish, resulting in an extreme low-resource setting (4k to 125k paired sentences), with challenges out as domain, dialectical, and orthographic mismatches between splits and datasets. The best systems shows that data cleaning and collection (\S\ref{sec:available_data}) as well as multilingual approaches (\S\ref{sec:mutlilingual_sup}) result in the best performance in this conditions. Finally the shared task on MT in Dravidian languages \cite{chakravarthi-etal-2021-findings-shared} features 3 languages paired with English as well as Tamil--Telugu. Again, the winning system uses a multilingual approach. The best performing systems use BT (\S\ref{subsec:backtranslation}) and BPE word segmentation (\S\ref{subsec:input_rep}).

The results from these challenges indicate that the optimal selection and combination of methods differs between cases (i.e., amount of monolingual, parallel data, cleanness of data, domain mismatch, linguistic closeness of languages). This implies that data analysis and linguistic knowledge are needed to improve a final system's performance.

\subsection{Transfer learning}

This helps low-resource tasks as a lower amount of data can be used for training. 
One application of transfer learning to MT is the usage of a pretrained RNN LM \cite{gulcehre2015using} as the decoder in an NMT system. \newcite{zoph2016transfer} 
is the first work that uses pretrained models to improve NMT systems. The authors perform two experiments with an RNN encoder--decoder architecture with an attention mechanism: the model is first pretrained on a high-resource language pair 
This works even better if related languages are used during pretraining
\cite{nguyen2017transfer}. Using pretrained LMs at decoding time and as priors at training time also improves vanilla models \cite{baziotis2020language}.

To avoid overfitting, models can be finetuned on both a HRLs pair and a LRLs pair in a multi-task fashion \cite{neubig2018rapid}.

However, how can we represent best the vocabulary?
\newcite{zoph2016transfer} use separate embeddings for the source and the target language. However, using tied embeddings has been shown to yield better results \cite{press2017using}.
\newcite{edunov2019pre} employs ELMO \cite{peters2018deep} 
representations as pretrained features in the encoder of a transformer model. 
\citet{song2020pre} shows that it is possible to improve performance by combining monolingual texts from linguistically related languages, performing a script mapping.  
It is also possible to extract features from a BERT model in the source language and combining these with an NMT system \cite{zhu2020incorporating}, but using a BERT model pretrained with a mixed sentences from source and target languages lead to even better results \cite{xu-etal-2021-bert}. 

Encoder-decoder pretrained models have gained popularity in the last years for low-resource MT. 
\newcite{conneau2019cross} proposes training the encoder and the decoder separately in order to get cross-language representations (XLM). 
This idea has further been extended by \citet[][ MASS]{song2019mass} to masking a \textit{sequence} of tokens from the input. 
Training MASS in a multilingual fashion and using monolingual data for pretraining helps to improve NMT for low-resource languages and zero-shot translation \cite{siddhant-etal-2020-leveraging}.
Another approach is to train the entire transformer model as a denoising autoencoder \citep[BART; ][]{lewis2019bart}.
The multilingual version of BART (mBART) is more suitable for NMT tasks and yields important gains \cite{liu2020multilingual}. It is also possible to pretrain a transformer in a multi-task, text-to-text fashion, where one of the tasks is MT \citep[T5;][]{raffel2020exploring}.
All four models can be finetuned for MT or used in an unsupervised fashion. Improvements to BART
can be obtained by augmenting the maximum likelihood objective with an additional objective, which is a data-dependent Gaussian prior distribution \cite{li2020data}. 
Huge LMs can improve zero-shot and few-shot learning even further \cite{brown2020language}, but at a high computational cost. 
Pursuing another direction, \newcite{wang2019denoising} develops a hybrid architecture between a transformer and a pointer-generator network.  At training time, the authors jointly train the encoder and the decoder in a denoising auto-encoding fashion. 

One crucial problem for transfer-learning is minimizing catastrophic forgetting \cite{serra2018overcoming}. \newcite{chen-etal-2021-zero} show that it is possible to combine a pre-trained multilingual model, with fine-tuining it with one single language pair, to improve zero-shot machine translation. Another way to handle this problem is reducing the number of parameter to be updated. \newcite{gheini-etal-2021-cross} propose to only update the cross attention parameters.

\subsection{Unsupervised MT}

The addition of other components such as masked LMs and denoising auto-encoding has also been tried \cite{stojanovski2019lmu}. Unsupervised methods are vulnerable to adversarial attacks of word substitution and order change in the input. Adversarial training can improve performance in such situations \cite{sun2020robust}. 
Since the initialization step is crucial for UMT, \newcite{ren-etal-2020-retrieve} aligns semantically similar  sentences from two monolingual corpora with the help of cross-lingual embeddings.
With these, an SMT system is trained to warm up an NMT system.
However, UMT still has to overcome a set of challenges. 
\newcite{sogaard2018limitations} shows that performance decays dramatically for languages with different typological features,  
since, in such situations, bilingual word embeddings \cite{conneau2017word} are far from isomorphic.
\newcite{vulic2020all} finds that isomorphism 
is also less likely if small amounts of monolingual data are used for training bilingual word embeddings. 
\newcite{nooralahzadeh2020zero} discovers that performance quickly deteriorates for a mismatch of source and target domain and that the initialization of word embeddings can affect MT performance. All of this makes  UMT for LRLs or endangered languages challenging. 

Some of the described issues have been addressed: 
\newcite{liu2019incorporating} proposes to combine word-level and subword-level embeddings to account for morphological complexity. 
For the problem of distant language pairs, \citet{leng2019unsupervised} proposes pivoting (cf. \S\ref{subsubsec:pivoting}). Isomorphism of bilingual word-embeddings can be improved with semi-supervised methods \cite{vulic-etal-2019-really}.

\citet{garcia2020multilingual} introduces multilingual UMT systems. The main idea consists of generalizing UMT 
by using a multi-way back-translation objective. 
Recently, pretrained multilingual transformer networks are used to improve UMT even further (cf.  \S\ref{subsec:transfer_learning}).

\section{Ethical Considerations}
Ethical concerns when working on MT for endangered languages 
include a lack of community involvement during language documentation, data creation, and development and setup of MT systems. For more information, we refer interested readers to \newcite{bird-2020-decolonising}.
Finally, we want to mention that publicly employing low-quality MT systems for LRLs bears a risk of translating incorrectly or in biased (e.g., sexist or racist) ways.

\end{document}